%% file: root.tex
\documentclass[a4paper, 10 pt, conference]{ieeeconf}

\usepackage[top=104pt, bottom=54pt, left=54pt, right=37pt, nohead, nofoot]{geometry}

\IEEEoverridecommandlockouts

\usepackage{dblfloatfix}

\usepackage{color, colortbl}
\usepackage[colorlinks,bookmarksnumbered,citecolor=orange,urlcolor=orange]{hyperref}
\usepackage{graphicx}
\graphicspath{{Figures}}
\DeclareGraphicsExtensions{.pdf,.png}
\usepackage{bigstrut}
\usepackage[english]{babel}

\usepackage{csquotes}

\usepackage[T1]{fontenc}
\usepackage{url}
\usepackage{multirow}
\usepackage{float}
\usepackage{xcolor}
\usepackage{hyperref}
 \hypersetup{
     colorlinks=true,
     linkcolor=orange,
     filecolor=orange,
     citecolor=orange,
     urlcolor=orange,
     }

\newcommand{\p}[1]{\small \noindent \textbf{{#1}.}}

\let\labelindent\relax
\usepackage{graphicx}
\usepackage{amsmath}
\usepackage{enumitem}

\usepackage{float}
\usepackage[ruled,vlined,linesnumbered]{algorithm2e}
\usepackage{epstopdf}
\usepackage{arydshln}

\usepackage{graphics} 
\usepackage{epsfig} 
\usepackage{times} 
\usepackage{amssymb}  
\usepackage{xcolor}

\usepackage{booktabs}
\usepackage{pifont}

\usepackage{cite}

\usepackage{multicol}
\usepackage{comment}
\usepackage{tikz}
\usetikzlibrary{arrows.meta,positioning,fit,backgrounds,shapes.multipart}
\usepackage{pgfplots}
\pgfplotsset{compat=1.17}
\usepgfplotslibrary{groupplots}

\usepackage[font={small}]{caption} 
\definecolor{keeprow}{RGB}{226,243,224}   
\definecolor{droprow}{RGB}{236,239,241}    
\newcommand{\sixpt}{\@setfontsize\sixpt{6}{8}}

\definecolor{cNaive}{HTML}{2A78D6}
\definecolor{cGraphRAG}{HTML}{EB6834}
\definecolor{cLightRAG}{HTML}{1BAF7A}
\definecolor{cEmbRAG}{HTML}{EDA100}
\definecolor{cOurs}{HTML}{E87BA4}
\definecolor{inkPrimary}{HTML}{0B0B0B}
\definecolor{inkMuted}{HTML}{898781}
\definecolor{gridLine}{HTML}{E1E0D9}

\definecolor{cCacheHit}{HTML}{2E8B57}
\definecolor{cCacheMiss}{HTML}{C0392B}

\definecolor{cCoffee}{HTML}{4A3AA7}
\definecolor{cDistractor}{HTML}{ADA995}

\title{\LARGE 
Embodied RAG++: Retrieval-Augmented Generation From What Exists to What Works in Embodied Scenes
}

\author{Fan Yang$^{1}$, Tatsuya Kikuzuki$^{1}$, Jun Liang$^{1}$, Jun Liu$^{2}$, Wenkai Li$^{2}$,\\ Ryosuke Kawamura$^{1}$, Ananya Hazarika$^{1}$, Prajit KrisshnaKumar$^{1}$, Shaunak Mehta$^{1}$,\\ and Yuushi Toyoda$^{1}$
\thanks{$^{1}$Fan Yang, Tatsuya Kikuzuki, Jun Liang, Ryosuke Kawamura, Ananya Hazarika, Prajit KrisshnaKumar, Shaunak Mehta, and Yuushi Toyoda are with Fujitsu. {\tt\small fan.yang@fujitsu.com}}%
\thanks{$^{2}$Jun Liu and Wenkai Li are with Carnegie Mellon University.}%
}
\begin{document}
\maketitle

\pagestyle{plain}
\thispagestyle{plain}

\begin{abstract} To serve a request in a human environment, a robot system must find an object that satisfies a functional requirement and a robot that can feasibly reach it. Scene graphs record which objects exist and where, but they often omit capabilities documented only in product manuals. Retrieval-augmented generation methods can read such manuals, yet they rarely reason jointly over object capabilities, physical instances, robot states, and dynamic connectivity. We introduce E2W-Bench, the Exists-to-Works Benchmark, which links dynamic hybrid scene graphs with real product manuals to evaluate capability-grounded robot--object assignment under availability, mobility, vertical-reach, and connector constraints. We also present Embodied RAG++, a spatially grounded retrieval framework that scopes retrieval to the manuals linked to candidate objects, caches extracted capabilities per product model, and resolves robot--object feasibility through deterministic spatial reasoning. On E2W-Bench, Embodied RAG++ reaches a pair accuracy of 0.90 with Qwen2.5-32B and 1.00 with Llama3.3-70B, compared with at most 0.30 and 0.55 for the baselines, while requiring the shortest preprocessing time among the retrieval-based methods and tying for the lowest total token count.
\end{abstract}

\section{Introduction} \label{sec:intro} 
Robots operating in human environments need more than knowledge of which objects exist and where they are located. To serve a request, a system has to find which physical object can satisfy the user's functional requirement and which available robot can feasibly reach it. Consider a request to heat a 27~L food container to 240\textdegree C. A scene graph may contain several microwave ovens, but the required cavity capacity and convection temperature are specified only in their product manuals. Even after identifying a suitable appliance, the system has to check whether an available robot can reach it under the current mobility and connectivity conditions. If an elevator becomes unavailable, for example, a wheeled robot may be unable to access another floor, whereas a legged robot may use the stairs. Resolving such a request therefore requires combining capabilities documented in manuals with the current state of the objects, robots, and connectors in the scene. 

Existing work addresses parts of this problem but not their combination. 3D scene graphs~\cite{armeni20193dscenegraph,hughes2022hydra}, open-vocabulary maps~\cite{gu2024conceptgraphs}, and hierarchical planners~\cite{rana2023sayplan} model spatial layouts and navigation paths, while benchmarks such as ScanRefer~\cite{chen2020scanrefer} and SQA3D~\cite{ma2023sqa3d} evaluate visual language grounding. These representations capture what exists and where, but not the non-visual functional attributes stored in external documentation. Graph-structured retrieval-augmented generation (RAG)~\cite{lewis2020rag} frameworks such as GraphRAG~\cite{edge2024graphrag}, RAPTOR~\cite{sarthi2024raptor}, and HippoRAG~\cite{gutierrez2024hipporag}, for their part, retrieve document knowledge but return textual answers that are not tied to physical instances and ignore reachability, robot availability, and changing scene state. EmbodiedRAG~\cite{xie2024embodiedrag} introduces spatial retrieval over observational memory, but its knowledge is limited to what the agent has perceived, not to external manuals. In addition, processing multi-page manuals anew for every query and every physical instance repeats the same work whenever several instances share a product model.

We introduce \textbf{E2W-Bench} (Exists-to-Works Benchmark), a benchmark for capability-grounded robot--object assignment over dynamic hybrid scene graphs. E2W-Bench connects physical object instances with product manuals and unifies robots, spatial connectors, and dynamic environmental states within a single graph. Each query has a machine-verified unique ground-truth pair, so a single exact-match check evaluates document reasoning and spatial allocation together.

\begin{figure}[t]
\centering
\includegraphics[width=\linewidth]{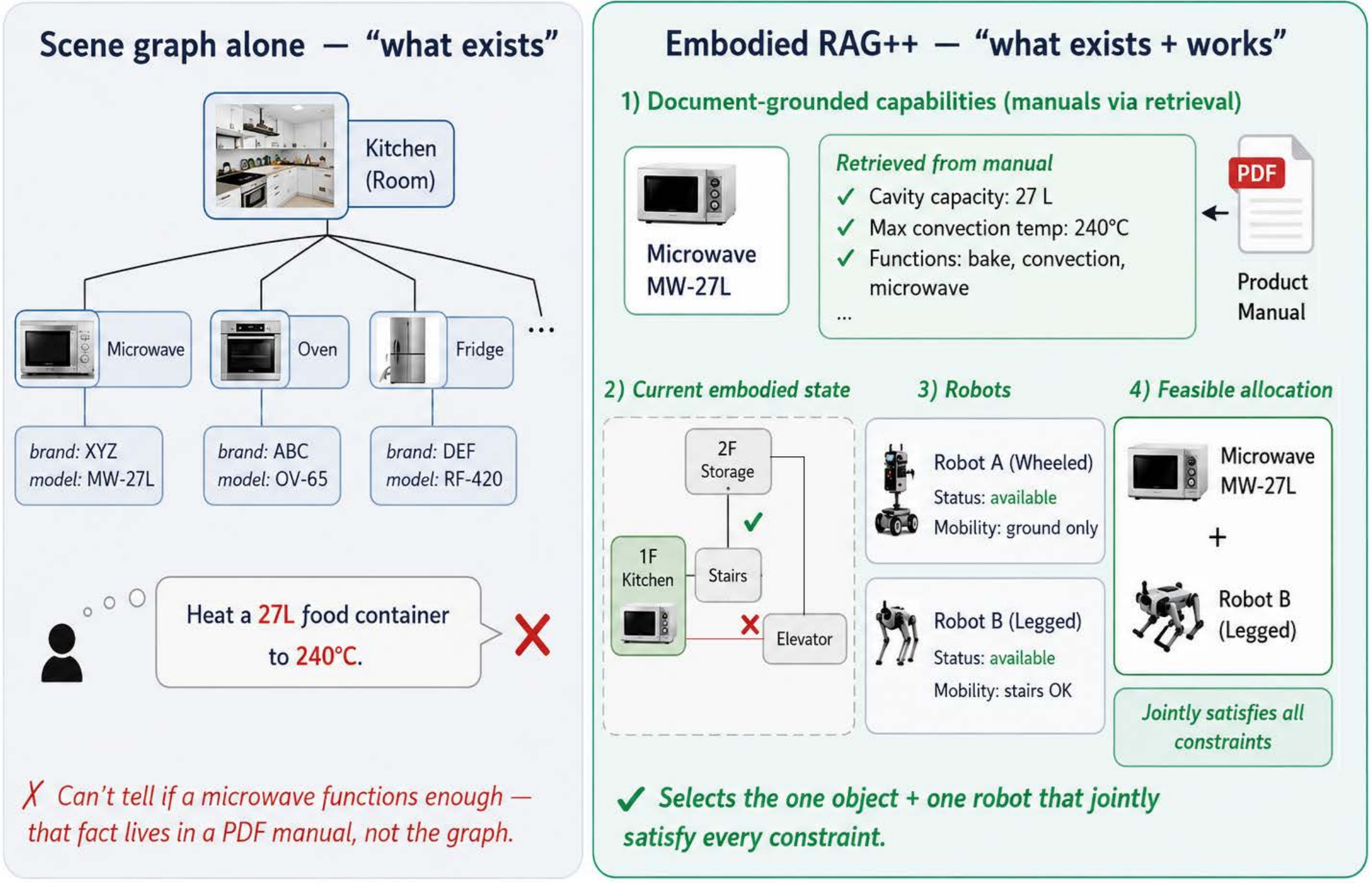}
\caption{Conventional scene graph versus Embodied RAG++. Left: a scene graph, represented by 3DSceneGraph~\cite{armeni20193dscenegraph}, records which appliances exist and their models, but cannot tell whether a microwave can heat a 27~L container to 240\textdegree C, because that fact is stated only in its product manual. Right: Embodied RAG++ retrieves the capability from the linked manual, checks the current connector states and robot mobility, and selects the object--robot pair that satisfies every constraint.}
\label{fig:demo}
\end{figure}

We also propose \textbf{Embodied RAG++}, a spatially grounded retrieval framework that couples document-grounded capability inference with deterministic feasibility verification (Fig.~\ref{fig:demo}). Embodied RAG++ uses the scene graph's object-to-manual links to restrict retrieval to the manuals of candidate objects. It resolves reachability, mobility, and connectivity constraints with deterministic spatial reasoning, which requires no large language model (LLM) calls. To avoid reprocessing the same manuals, Embodied RAG++ stores extracted facts in \emph{Capability Cards}, memory units keyed by product model that grow incrementally as queries request new fields. Every physical instance of a model and every later query that needs the same field reuses the card.

Our contributions are as follows:
\begin{itemize}[leftmargin=*] \item \textbf{E2W-Bench}, which links dynamic hybrid scene graphs to real product manuals and gives every query a machine-verified unique ground-truth pair (\S\ref{sec:benchmark}). \item \textbf{Embodied RAG++}, which combines manual-scoped, aspect-split retrieval, model-level Capability Cards, availability filtering, and deterministic spatial reasoning (\S\ref{sec:method}). \item An evaluation with three LLM backbones and four baselines, in which Embodied RAG++ reaches a pair accuracy of 0.90 with Qwen2.5-32B and 1.00 with Llama3.3-70B, compared with at most 0.30 and 0.55 for the baselines, with the shortest preprocessing time among the retrieval-based methods and the lowest total token count, tied with EmbodiedRAG (\S\ref{sec:experiments}). \end{itemize}

\section{Related Work} \label{sec:related} 

\subsection{Scene Graphs and Embodied Scene Representations} Scene graphs represent environments as objects, rooms, floors, and their spatial and semantic relations~\cite{armeni20193dscenegraph}. Recent systems construct these representations online through hierarchical metric-semantic mapping~\cite{hughes2022hydra}, incorporate open-vocabulary concepts from vision-language models~\cite{gu2024conceptgraphs}, and support language-conditioned planning over large environments~\cite{rana2023sayplan}. Related benchmarks evaluate grounding free-form language to objects in 3D scenes~\cite{chen2020scanrefer} and answering questions from an agent's situated viewpoint~\cite{ma2023sqa3d}. These methods capture the spatial and semantic structure of a scene, but they do not recover functional capabilities stated only in external documents such as product manuals, nor do they determine whether an available robot can reach and use a qualifying object as the scene changes. 

\subsection{Retrieval-Augmented Generation over Structured Corpora} Retrieval-augmented generation (RAG) grounds language-model outputs in external documents~\cite{lewis2020rag}. Structured variants organize document collections to improve retrieval and multi-hop reasoning. GraphRAG constructs an entity-relation graph and summarizes its communities~\cite{edge2024graphrag}; RAPTOR recursively clusters and summarizes passages into a retrieval tree~\cite{sarthi2024raptor}; and HippoRAG applies personalized PageRank over a knowledge graph to integrate information across passages~\cite{gutierrez2024hipporag}. These methods improve document-grounded question answering, but their preprocessing generally spans the corpus and their outputs remain passages or textual responses. In our setting, retrieved evidence must instead be grounded to a physical object instance and checked against the current state of robots and connectors to produce an executable robot--object assignment. 

\subsection{Embodied Retrieval-Augmented Generation} 

EmbodiedRAG constructs a hierarchical, non-parametric memory from an agent's observations and retrieves multi-resolution semantic and spatial information for navigation and language generation~\cite{xie2024embodiedrag}. OpenSG-RAG~\cite{fei2026open} encodes open-world 3D scene graphs in a vector database to support retrieval-augmented scene question answering, visual grounding, instance retrieval, and task planning. RAG-3DSG~\cite{chang2026rag} instead applies object-level retrieval during scene-graph construction, using semantically reliable objects as contextual anchors to refine uncertain object predictions. These methods retrieve from observed scene information or use retrieval to improve the scene representation itself. Embodied RAG++ instead treats the scene graph as an index into external knowledge: each physical object instance links to a product manual that states capabilities a camera often cannot observe. It reuses capabilities extracted per product model and checks the selected object against the current robot and connector states to produce an executable robot--object assignment.

\section{E2W-Bench}
\label{sec:benchmark}

\input{figures/benchmark_scene_graph}

\subsection{Task Formulation}
\label{sec:task_formulation}
We formulate E2W-Bench as a robot--object pair selection problem over a \emph{dynamic hybrid scene graph} with four node/edge types: \emph{appliance nodes} (category, model, a link to that model's real product manual, availability, position); \emph{robot nodes} (mobility profile, availability, position); a \emph{spatial hierarchy} of scene $\rightarrow$ floor $\rightarrow$ room $\rightarrow$ object/robot; and \emph{connector edges} (stairs, elevators, doors) carrying a traversal cost, an accessibility list of robot types, and a status that can change between queries. Fig.~\ref{fig:benchmark_scene_graph} shows one such graph in full. Its node and edge states are not fixed at construction time; they track the physical scene and change as a robot moves or is dispatched, an appliance is switched on or taken out of service, or a connector's status changes. These state updates make the graph \emph{dynamic} rather than a static record of object positions and labels.

Given the graph's current state and a natural-language query, a system returns a robot identifier, a target-object identifier, and both of their positions. A returned pair is correct only if four independent conditions hold: (i) \emph{capability}: the target's real manual establishes that it meets the query's functional requirement; (ii) \emph{availability}: that instance is not \texttt{in\_use} or \texttt{out\_of\_order}; (iii) \emph{mobility}: a candidate robot can reach the target's floor given the current connector states and its own accessibility profile; and (iv) \emph{vertical reach}: that robot's body or arm can physically reach the object's mounted height.

\subsection{Archetype-Based Generation}
\label{sec:archetype_generation}

We generate benchmark scenes in the MuJoCo simulator using three robot platforms: a quadruped, a humanoid, and a wheeled mobile manipulator equipped with an AgileX Piper arm. The scenes include appliance 3D models reconstructed from real objects and their corresponding product manuals. Fig.~\ref{fig:dataset_stats} summarizes the 20 released scenes and the underlying manual corpus. Each scene is a verified snapshot of the dynamic graph defined in \S\ref{sec:task_formulation}: it fixes robot positions, appliance availability, and connector states, the same fields a deployed system updates online, rather than replaying continuous physical dynamics.

Each scene instantiates one of 10 hand-designed \emph{query archetypes}, with two archetypes for each of five appliance categories: microwave, rice cooker, washing machine, water dispenser, and bread maker. Each archetype has two scene variations, yielding 20 scenes in total. A \emph{threshold} archetype requires the target to satisfy a compound predicate, such as cavity capacity $\geq 27$~L and bake temperature $\geq 240$\textdegree C, while every listed distractor fails at least one condition. An \emph{extremum} archetype requires all candidates to satisfy a baseline predicate, with the target minimizing or maximizing a specified field. Before scene generation, we manually extract the relevant capability values from the 13 product manuals covering these five categories (Fig.~\ref{fig:dataset_stats}b) and verify each predicate against the resulting reference table, so every scene agrees with its source manuals. The reference table is used only to construct and validate the ground truth; at inference time, Embodied RAG++ derives its Capability Cards directly from the manuals (\S\ref{sec:capability_cards}).

Each scene contains the qualifying appliance, all distractors specified by the archetype, and several unrelated manual-linked appliances and furniture items. These objects are distributed across one to four floors connected by parallel elevators and stairs. Each scene also includes two to seven robots of three types, using their published mobility and vertical-reach envelopes. Two independent \emph{reachability traps} make incorrect answers appear plausible when either constraint is ignored. The \emph{mobility trap} places an otherwise closer, stairs-incapable wheeled robot behind a connector whose elevator is out of service. The \emph{height trap} places roughly half of the normally counter-height targets, including microwaves, rice cookers, and bread makers, on shelves reachable only within a humanoid's extended vertical envelope, while an otherwise suitable wheeled or quadruped robot is positioned nearby. Because the two traps are independent, a scene may contain a mobility trap, a height trap, both, or neither.

\input{figures/dataset_stats}

\subsection{Evaluation Protocol}
\label{sec:evaluation_protocol}

E2W-Bench evaluates a system along five axes that distinguish \emph{getting the right answer}, \emph{getting it for the right reasons}, and \emph{getting it efficiently}. \textbf{Functional correctness} is exact-match accuracy on the returned \texttt{(robot\_id, target\_object\_id)} pair. \textbf{Robot--object feasibility} requires the selected pair to be jointly reachable under the current connector and mobility states, and \textbf{position-level grounding} requires the returned coordinates to match the recorded positions of the pair. \textbf{Evidence grounding} checks whether the qualifying capability is supported by the relevant manual span or Capability Card field. \textbf{Retrieval efficiency} measures how much manual text is read and how often lookups reuse Capability Cards instead of triggering new extraction. Because Capability Cards cache facts rather than final decisions (\S\ref{sec:capability_cards}), retrieval efficiency is independent of correctness and is evaluated separately through cache behavior (\S\ref{sec:cache_results}).

Functional correctness, feasibility, and position-level grounding can all be evaluated from the returned pair. The ground truth of each sample is a unique pair verified by brute-force search, and every distractor and reachability trap described in \S\ref{sec:archetype_generation}, including unmet capability predicates, unavailable instances, mobility traps, and height traps, leads to a specific incorrect pair. A system can return the correct pair only if all four conditions in \S\ref{sec:task_formulation} hold, and the positions of a correct pair are fixed by the graph. Exact-match pair accuracy in \S\ref{sec:experiments} is thus a joint measure of these three axes.

Pair correctness does not certify evidence grounding. A system may return the correct pair while citing an incorrect manual passage, relying on an unsupported field, or providing no evidence: the traps detect reasoning failures that change the final answer, not unsupported reasoning that happens to reach it. Capability Cards retain the source manual span for every stored field (\S\ref{sec:capability_cards}), so evidence grounding can be audited, but this paper does not report a quantitative grounding score.

Each E2W-Bench scene is far denser than the scenes of existing benchmarks (Table~\ref{tab:benchmark_comparison}). Its serialized spatial context reaches approximately 500K tokens, compared with fewer than 10K for 3DSceneGraph, ScanQA, and SQA3D, and it is paired with 770 pages of real product manuals. Each scene also contains two to seven robots and dynamic connector and appliance-availability states, so every query must be answered against a large, interdependent context.

\begin{table}[t] 
  \centering \caption{Comparison with representative scene-based benchmarks.} \label{tab:benchmark_comparison} \resizebox{\columnwidth}{!}{ \begin{tabular}{lcccc} \toprule \textbf{Benchmark} & \textbf{\# Scenes} & \textbf{Scene State} & \textbf{Robots per Scene} & \textbf{Scene Tokens} \\ \midrule 3DSceneGraph~\cite{armeni20193dscenegraph} & 572 & Static & 1 & $<10$K \\ ScanQA~\cite{azuma2022scanqa} & \textbf{800} & Static & 1 & $<10$K \\ SQA3D~\cite{ma2023sqa3d} & 650 & Static & 1 & $<10$K \\ \textbf{E2W-Bench (ours)} & 20 & Dynamic & \textbf{2--7} & $\approx\textbf{500}$K \\ \bottomrule \end{tabular} } 
\end{table}

\section{Embodied RAG++ Method}
\label{sec:method}

Embodied RAG++ answers a query in two coupled stages instead of running retrieval and spatial reasoning as independent subsystems. The first stage determines which candidate objects satisfy the functional requirement: it retrieves only from the manuals linked to the candidates, reuses capability fields already extracted for the same product model, and judges each model against the requirement. The second stage passes the qualifying object to a deterministic spatial layer that selects a feasible robot. We describe each stage in turn; Algorithm~\ref{alg:pipeline} gives the full procedure.

\subsection{Problem Setup}
Let the scene graph provide, for every appliance instance $o$, a category $\mathrm{cat}(o)$, a model identifier $\mathrm{model}(o)$, a link to a manual document $\mathrm{manual}(o)$, an availability state, and a position; and for every robot $r$, an availability state, a mobility profile, and a position. Given a query $q$, Embodied RAG++ must return a pair $(r^\star, o^\star)$ such that $o^\star$ satisfies $q$'s functional requirement on manual evidence, $o^\star$ is available, and $r^\star$ is available and feasible for $o^\star$ under the current connector graph.

\subsection{Capability Cards}
\label{sec:capability_cards}
Product manuals are long, multi-topic documents, and re-reading them for every query does not scale with the number of instances or queries. Embodied RAG++ therefore treats each manual as the authoritative source and maintains a \emph{Capability Card}: a per-model cache containing only previously requested capability fields and their supporting manual spans. Cards are keyed by \emph{model identifier}, allowing all instances of the same model to share evidence. Nothing is extracted in advance. A card is created when a model is first queried and stores only the fields required by that query, avoiding an offline corpus pass that would need to anticipate all capabilities future queries might request. If a required field is cached, it is reused without retrieval or extraction. Otherwise, a \emph{refine} step retrieves the missing aspect from the linked manual (\S\ref{sec:scoped_retrieval}), extracts it, and adds it to the existing card. Card coverage grows incrementally, and a card is never rebuilt. Because cards store capability fields rather than query-specific verdicts, cached and newly extracted fields are judged identically against the current query, so caching reduces retrieval and generation cost without changing the judgment. When extraction is required, each manual excerpt is processed separately to reduce the risk of mixing values from different operating modes.

\subsection{Scoped, Aspect-Split Retrieval}
\label{sec:scoped_retrieval}
A global top-$k$ search over the whole manual corpus ranks passages only by their similarity to the query wording. The highest-ranked passage may therefore come from a short manual that happens to use similar phrasing rather than from the long, differently worded manual that actually specifies the candidate. Embodied RAG++ avoids this by using information the scene graph already provides: the manual $\mathrm{manual}(o)$ linked to each candidate object $o$. The query is embedded once, a wide candidate pool is retrieved from a single index over the entire corpus, and the pool is filtered to passages whose source document is $\mathrm{manual}(o)$. One corpus-wide index therefore suffices, and every retained passage comes from the candidate's own manual. Compound requirements are further split into per-aspect queries. For example, ``holds at least 10~kg'' and ``supports a 24-hour delay start'' are retrieved separately, because in a single combined query the wording of one aspect can dominate the embedding and push the other aspect's supporting passage out of the retrieved pool. Together, scoping and aspect splitting ensure that every generation call in Embodied RAG++ operates on a few short passages or on a small dictionary of previously extracted fields, never on a full manual, which keeps each call short and traceable to its source.

This design also distinguishes Embodied RAG++ from document-first graph construction such as GraphRAG~\cite{edge2024graphrag}. GraphRAG calls an LLM to extract entities and relations from every chunk of every document before it can answer any query, so its one-time cost grows with the size of the corpus. Embodied RAG++ builds no corpus-wide index of extracted facts; it extracts a field only when a query first requests it, so its generation cost grows with the number of distinct capability fields queried rather than with the number of manuals or chunks in the corpus.

\subsection{Graph-Based Selective-Recomputation Retrieval Backbone}
Every scoped search in \S\ref{sec:scoped_retrieval} runs on a pruned proximity graph over the corpus rather than on a stored embedding table. Each chunk is a node of a hierarchical navigable small-world graph~\cite{malkov2020hnsw} built with a graph-index backend~\cite{johnson2019faiss}, and high-degree-preserving pruning keeps only the edges needed for long-range navigation. Only the graph topology is stored; the chunk embeddings used to build it are discarded~\cite{wang2025leann}. At query time, best-first beam search with beam width $\mathrm{ef}$ traverses the graph, and only the embeddings of the visited chunks are recomputed with a local dense encoder~\cite{izacard2022contriever}. The storage cost of the corpus-wide index is therefore that of a graph rather than an embedding matrix.

Embodied RAG++ keeps one such graph for the entire manual corpus rather than one per model or per manual, so adding a new appliance model to a deployment does not require building a new index. Scoping is applied after the graph search rather than by restricting it: each query uses a wider beam and a larger candidate count than plain top-$k$ search, and the retrieved pool is filtered to passages from the candidate's $\mathrm{manual}(o)$ before the top-$k$ are kept. Each aspect-split query runs its own search-and-filter pass over the same graph.

The backbone is invoked only on a Capability Card miss. When the required field is already on the model's card, \textsc{Judge} (Algorithm~\ref{alg:pipeline}) reads it directly, with no query embedding, graph search, or embedding recomputation. Retrieval for a given model and capability field therefore runs at most once across a deployment's query history, regardless of how many instances of that model exist or how many later queries request the same field.

\subsection{Spatial Reasoning}
\label{sec:spatial_reasoning}
Once a qualifying object $o^\star$ has been identified from Capability Card evidence, feasibility is resolved by deterministic operators over fields the scene graph already stores, without any LLM call. \textsc{FeasibleRobots} applies two checks to every candidate robot. \textsc{Reachable} runs a breadth-first search over the floor--connector graph and traverses an edge only if the connector is operational or clear and the robot's type appears in the connector's accessibility list; an elevator outage or a robot without stair-climbing capability therefore removes the corresponding path. \textsc{HeightReachable} compares the target's recorded mounting height with a fixed vertical-reach envelope for each robot type, derived from the platform's published arm and body dimensions, and excludes robots that can navigate to the target but cannot physically reach it. \textsc{FeasibleRobots} keeps the available robots that pass both checks, and \textsc{PickNearest} breaks any remaining tie by shortest-path navigation distance. Because this layer only reads the graph, it cannot return a robot identifier absent from the graph or ignore a connector's current status. \textsc{Reachable}, \textsc{HeightReachable}, and \textsc{PickNearest} apply the same reachability, height, availability, and tie-breaking rules used to construct and validate the benchmark ground truth (\S\ref{sec:archetype_generation}).

\begin{algorithm}[t]
\SetAlgoLined
\footnotesize
\KwIn{Dynamic hybrid scene graph $G$; query $q$; Capability Card store $\mathrm{Cards}$}
\KwOut{Feasible pair $(r^\star, o^\star)$ with positions}
$(\mathrm{cat}, \mathrm{mode}, \phi) \leftarrow \textsc{Classify}(q)$\;
$\mathcal{O} \leftarrow \{\, o \in G : \mathrm{cat}(o) = \mathrm{cat} \,\}$\;
$\mathcal{M} \leftarrow \{\, \mathrm{model}(o) : o \in \mathcal{O} \,\}$\;
\ForEach{$m \in \mathcal{M}$}{
  $o_m \leftarrow$ any $o \in \mathcal{O}$ with $\mathrm{model}(o) = m$\;
  $c_m \leftarrow \mathrm{Cards.get}(m, \emptyset)$\;
  $F_{\mathrm{miss}} \leftarrow \mathrm{FIELDS}(\phi) \setminus \mathrm{COVERAGE}(c_m)$\;
  \If{$F_{\mathrm{miss}} \neq \emptyset$}{
    $P \leftarrow \textsc{ScopedRetrieve}(F_{\mathrm{miss}}, \mathrm{manual}(o_m))$\;
    $\Delta c_m \leftarrow \textsc{Extract}(P, F_{\mathrm{miss}})$\;
    $c_m \leftarrow \textsc{Merge}(c_m, \Delta c_m)$;\quad $\mathrm{Cards}[m] \leftarrow c_m$\;
  }
  $v_m \leftarrow \textsc{Judge}(c_m, \phi)$\;
}
$\mathcal{O}^\star \leftarrow \{\, o \in \mathcal{O} : v_{\mathrm{model}(o)} = \textsc{Pass} \text{ and available}(o) \,\}$\;
\If{$\mathrm{mode} = \textsc{extremum}$ \textnormal{\textbf{and}} $|\mathcal{O}^\star| > 1$}{
  $\mathcal{O}^\star \leftarrow \{\, \textsc{ResolveExtremum}(\mathcal{O}^\star, \phi) \,\}$\;
}
$o^\star \leftarrow$ the unique element of $\mathcal{O}^\star$\;
$\mathcal{R}^\star \leftarrow \textsc{FeasibleRobots}(G, o^\star)$\;
$r^\star \leftarrow \textsc{PickNearest}(\mathcal{R}^\star, o^\star)$\;
\Return{$(r^\star, o^\star)$ with their recorded positions}
\caption{Embodied RAG++ Pipeline}
\label{alg:pipeline}
\end{algorithm}

\section{Experiments}
\label{sec:experiments}

\begin{table*}[t] 
\centering 
\caption{Exact-match accuracy of robot selection (Robot), object selection (Object), and the joint robot--object pair (Pair) on the 20 E2W-Bench queries, one per scene, across systems and LLM backbones. ``+ SR'' denotes use of the shared deterministic Spatial Reasoning module (\S\ref{sec:spatial_reasoning}). Naive LLM is reported both without and with this module to isolate its contribution; all other baselines use the same module as Embodied RAG++.} \label{tab:answer_exact_match}
\footnotesize
\begin{tabular}{lccc ccc ccc} \toprule \multirow{2}{*}{System} & \multicolumn{3}{c}{Qwen2.5-32B} & \multicolumn{3}{c}{Llama3.3-70B} & \multicolumn{3}{c}{Gemini-3.5-Flash} \\ \cmidrule(lr){2-4} \cmidrule(lr){5-7} \cmidrule(lr){8-10} & Robot & Object & Pair & Robot & Object & Pair & Robot & Object & Pair \\
\midrule
 Naive LLM& 0.20& 0.45 & 0.10& 0.45& 0.65 & 0.30& 0.50& 0.95 &0.50\\  
 Naive LLM + SR & 0.40 & 0.45 & 0.20 & 0.70 & 0.65 & 0.35 & 1.00 & 0.95 & 0.95 \\ GraphRAG + SR & 0.55 & 0.45 & 0.30 & 0.75 & 0.70 & 0.55 & 1.00 & 0.95 & 0.95 \\ LightRAG + SR & 0.55 & 0.45 & 0.25 & 0.35 & 0.45 & 0.35 & 1.00 & 0.85 & 0.85 \\ EmbodiedRAG + SR & 0.60 & 0.45 & 0.30 & 0.60 & 0.75 & 0.50 & 1.00 & 0.95 & 0.95 \\ \textbf{Embodied RAG++ (ours)} & \textbf{0.90} & \textbf{0.90} & \textbf{0.90} & \textbf{1.00} & \textbf{1.00} & \textbf{1.00} & \textbf{1.00} & \textbf{1.00} & \textbf{1.00} \\ \bottomrule
\end{tabular}
\end{table*}

\input{figures/efficiency_comparison}

\input{figures/cache_hit_rate}

\begin{table}[th!] \centering \caption{Ablation study of Embodied RAG++ on the 20 E2W-Bench queries. Each row removes exactly one component; we report exact-match accuracy of robot selection (Robot), object selection (Object), and the joint pair (Pair).} \label{tab:ablation}
\setlength{\tabcolsep}{2.5pt} 
\footnotesize
\begin{tabular}{lccc ccc} \toprule \multirow{2}{*}{Configuration} & \multicolumn{3}{c}{Qwen2.5-32B} & \multicolumn{3}{c}{Llama3.3-70B} \\ \cmidrule(lr){2-4} \cmidrule(lr){5-7} & Robot & Object & Pair & Robot & Object & Pair \\ \midrule \textbf{Embodied RAG++ (full)} & \textbf{0.90} & \textbf{0.90} & \textbf{0.90} & \textbf{1.00} & \textbf{1.00} & \textbf{1.00} \\ \midrule
   \quad w/o Capability Cards & 0.90 & 0.90 & 0.90 & 1.00 & 1.00 & 1.00 \\ 
  \quad w/o Model-level Sharing & 0.85 & 0.85 & 0.85 & 0.75 & 0.75 & 0.75 \\
  \quad w/o Aspect Splitting & 0.80 & 0.75 & 0.75 & 0.65 & 0.65 & 0.65 \\ \quad w/o Spatial Reasoning & 0.65 & 0.85 & 0.65 & 0.45 & 0.75 & 0.45 \\ \bottomrule \end{tabular}
\end{table}

We evaluate Embodied RAG++ against four baselines on all 20 released E2W-Bench scenes, one query per scene, with three LLM backbones. Table~\ref{tab:answer_exact_match} reports exact-match accuracy on the returned \texttt{(robot\_id, target\_object\_id)} pair, which jointly measures functional correctness, robot--object feasibility, and position-level grounding, but not evidence grounding or retrieval efficiency (\S\ref{sec:evaluation_protocol}). We further analyze the hit-rate behavior of the Capability Card cache (\S\ref{sec:cache_results}) and ablate each component of Embodied RAG++ (\S\ref{sec:ablation}).

\subsection{Baselines}
\label{sec:baselines}
Every baseline except Naive LLM shares the deterministic Spatial Reasoning (SR) module (\S\ref{sec:spatial_reasoning}): each baseline identifies only the qualifying object $o^\star$, and the module selects the robot for it, exactly as in Embodied RAG++. Spatial reasoning, which applies the same rules that generate the benchmark ground truth, is therefore held constant across Table~\ref{tab:answer_exact_match}, and the comparison isolates the capability-retrieval architecture. Every method runs on the same LLM backbones: \texttt{qwen2.5:32b}~\cite{qwen2024qwen25} and \texttt{llama3.3:70b}~\cite{grattafiori2024llama3} served locally via Ollama, and \texttt{gemini-3.5-flash}~\cite{googledeepmind2026gemini35flash}.

\p{Naive LLM} One LLM call per query over the complete scene graph and every candidate manual's full text, with no retrieval, chunking, or ranking; the same call also proposes the robot, so feasibility is resolved by the LLM itself. This baseline tests whether full-context prompting solves the task unaided, and at what token cost.

\p{Naive LLM + SR} The same call, restricted to proposing only $o^\star$; the SR module selects the robot. Comparing it with Naive LLM isolates the contribution of the SR module.

\p{GraphRAG + SR}~\cite{edge2024graphrag} Uses GraphRAG's default entity/relationship extraction pipeline over the same inputs (sanitized scene graph, height constraints, full manual text) to identify $o^\star$; one-time construction cost is reported separately from per-query cost.

\p{LightRAG + SR}~\cite{guo2024lightrag} Builds its own entity--relation graph from the manual corpus via LLM extraction to identify $o^\star$; each sample's scene graph is inserted as a transient document before querying and removed after, so scene state never leaks across scenes.

\p{EmbodiedRAG + SR}~\cite{xie2024embodiedrag} EmbodiedRAG has no native document support, so we adapt only its graph-retrieval component: the scene graph becomes its hierarchical semantic-forest form for scene-node retrieval, paired with a separate dense document retriever for manual passages, to identify $o^\star$. This baseline tests whether independent spatial and document retrieval suffices for joint reasoning.

\subsection{Results}

The open-weight backbones separate the methods most clearly (Table~\ref{tab:answer_exact_match}). With Qwen2.5-32B, Embodied RAG++ reaches 0.90 pair accuracy, 60~percentage points (pp) above the strongest baselines, GraphRAG and EmbodiedRAG (0.30). With Llama3.3-70B, it reaches 1.00, 45~pp above GraphRAG (0.55), and it is the only evaluated system with perfect pair accuracy on an open-weight backbone. With Gemini-3.5-Flash, most systems approach the ceiling: Embodied RAG++ reaches 1.00, three + SR baselines reach 0.95, LightRAG + SR reaches 0.85, and Naive LLM without SR reaches 0.50.

Fig.~\ref{fig:efficiency_comparison} compares efficiency. GraphRAG and LightRAG require 3--12~h of index construction; GraphRAG then answers each query in 6--12~s, whereas LightRAG is the slowest at 46--70~s. EmbodiedRAG preprocesses in 43--46~s and answers fastest, in 7--10~s. Embodied RAG++ preprocesses in 14--25~s, the shortest among the retrieval-based methods, answers in 15--20~s, faster than Naive LLM and LightRAG, and ties with EmbodiedRAG for the lowest total token count (0.2M tokens over the evaluation). After the first queries, most of its capability lookups are served from Capability Cards without new retrieval or extraction (\S\ref{sec:cache_results}).

Because all baselines use the same SR module, the accuracy differences between them and Embodied RAG++ do not come from feasibility resolution. Adding the module to Naive LLM raises robot and pair accuracy on every backbone while leaving object accuracy unchanged, but Naive LLM + SR remains far below Embodied RAG++ on the open-weight backbones. The ablations in Table~\ref{tab:ablation} locate the remaining gap in retrieval design: removing aspect splitting or model-level sharing reduces accuracy on both backbones, whereas replacing Capability Cards with raw retrieved excerpts does not.

\subsection{Capability Card Cache Behavior}
\label{sec:cache_results}
To characterize the Capability Card cache, we replay the 20-query evaluation log in order from a cold cache and record, for every judged candidate, whether its model's card already covered the required fields (a hit) or required new extraction (a miss). Fig.~\ref{fig:cache_hit_rate}(a) shows that misses concentrate in the first queries, while models are still unseen; after a model's first query, every later query that involves that model reads the required capability from its card. In Fig.~\ref{fig:cache_hit_rate}(b), the cumulative hit rate rises throughout the run, and the steady-state hit rate measured over a full deployment history exceeds 85\%: most capability lookups in deployment require no new retrieval or extraction call. The baselines have no counterpart to this curve. GraphRAG and LightRAG pay their extraction cost upfront over the whole corpus, regardless of which facts later queries need, while Naive LLM and EmbodiedRAG keep no model-keyed memory, so their per-query cost does not decrease as queries about the same appliance model accumulate.

\subsection{Ablation Study}
\label{sec:ablation}
Table~\ref{tab:ablation} isolates the contribution of each component by removing one at a time on the same 20 E2W-Bench scenes with the two open-weight backbones. Gemini-3.5-Flash is excluded because the full method already reaches near-ceiling accuracy with it (Table~\ref{tab:answer_exact_match}), which leaves little room to measure degradation.

\textbf{Capability Cards} do not affect accuracy. Replacing them with raw retrieved excerpts, so that \textsc{Judge} reasons directly over the retrieved passages instead of cached fields, yields identical accuracy on both backbones. Accuracy therefore depends on the extracted content, not on whether it is cached. The role of the cards is to avoid re-retrieving and re-extracting the same fields for later instances and queries (\S\ref{sec:cache_results}).

Removing \textbf{model-level sharing}, that is, keeping a separate card for each physical instance instead of one card per model, lowers pair accuracy by 5~pp on Qwen2.5-32B and by 25~pp on Llama3.3-70B, and requires roughly six to seven times more LLM calls. This variant still extracts and caches fields, but once per instance rather than once per model. Each instance therefore requires an independent extraction from the manual, and each extraction is another opportunity for the LLM to misread it, which affects Llama3.3-70B more. Sharing cards across the instances of a model thus reduces both extraction errors and the number of LLM calls.

Removing \textbf{aspect splitting} lowers object accuracy from 0.90 to 0.75 on Qwen2.5-32B and from 1.00 to 0.65 on Llama3.3-70B. With a single embedding for a compound requirement, one constraint can dominate the query and push the evidence for the other out of the retrieved pool (\S\ref{sec:scoped_retrieval}).

Removing \textbf{spatial reasoning} causes the largest drop. Object accuracy decreases moderately, to 0.85 and 0.75, while robot accuracy falls to 0.65 on Qwen2.5-32B and 0.45 on Llama3.3-70B. No baseline corresponds to this row, because all baselines keep the SR module (\S\ref{sec:baselines}). The drop shows that the LLM backbones do not reliably reproduce reachability and height reasoning from a serialized scene graph, which the deterministic layer performs exactly.

Both backbones show the same ordering: spatial reasoning has the largest effect on accuracy, followed by aspect splitting and model-level sharing, while Capability Cards have no measurable effect. Llama3.3-70B degrades more than Qwen2.5-32B under most ablations, so the larger backbone does not remove the dependence on these components.

\vspace{0.5em}
\section{Conclusion}
\label{sec:conclusion}
This paper addressed capability-grounded robot--object assignment: identifying which physical object satisfies a functional requirement documented only in an external manual, and which available robot can feasibly reach it under dynamic scene conditions. We introduced E2W-Bench, which links dynamic hybrid scene graphs to real product manuals with machine-verified unique ground truth, and Embodied RAG++, which scopes retrieval to the linked manual, caches extracted capabilities per product model in reusable Capability Cards, and delegates feasibility to deterministic spatial reasoning. Across three LLM backbones and four baselines, all but one of which share the same spatial reasoning module, Embodied RAG++ achieves the highest pair accuracy and ties EmbodiedRAG for the lowest total token count. The ablations attribute its advantage over the baselines to aspect-split retrieval and model-level sharing of Capability Cards, and show that removing deterministic spatial reasoning causes the largest accuracy drop. Our evaluation assumes that each object's link to its manual is known and is conducted in simulation. Future work will broaden E2W-Bench's category and archetype diversity, infer the object-to-manual link from visual or textual evidence, and validate the spatial-reasoning layer on physical robots under real perception and localization noise.


\vspace{0.5em}
\bibliographystyle{IEEEtran}
\bibliography{IEEEabrv,bibtex}

\end{document}

%% file: figures/benchmark_scene_graph.tex
\begin{figure*}[t]
\centering
\includegraphics[width=0.97\textwidth]{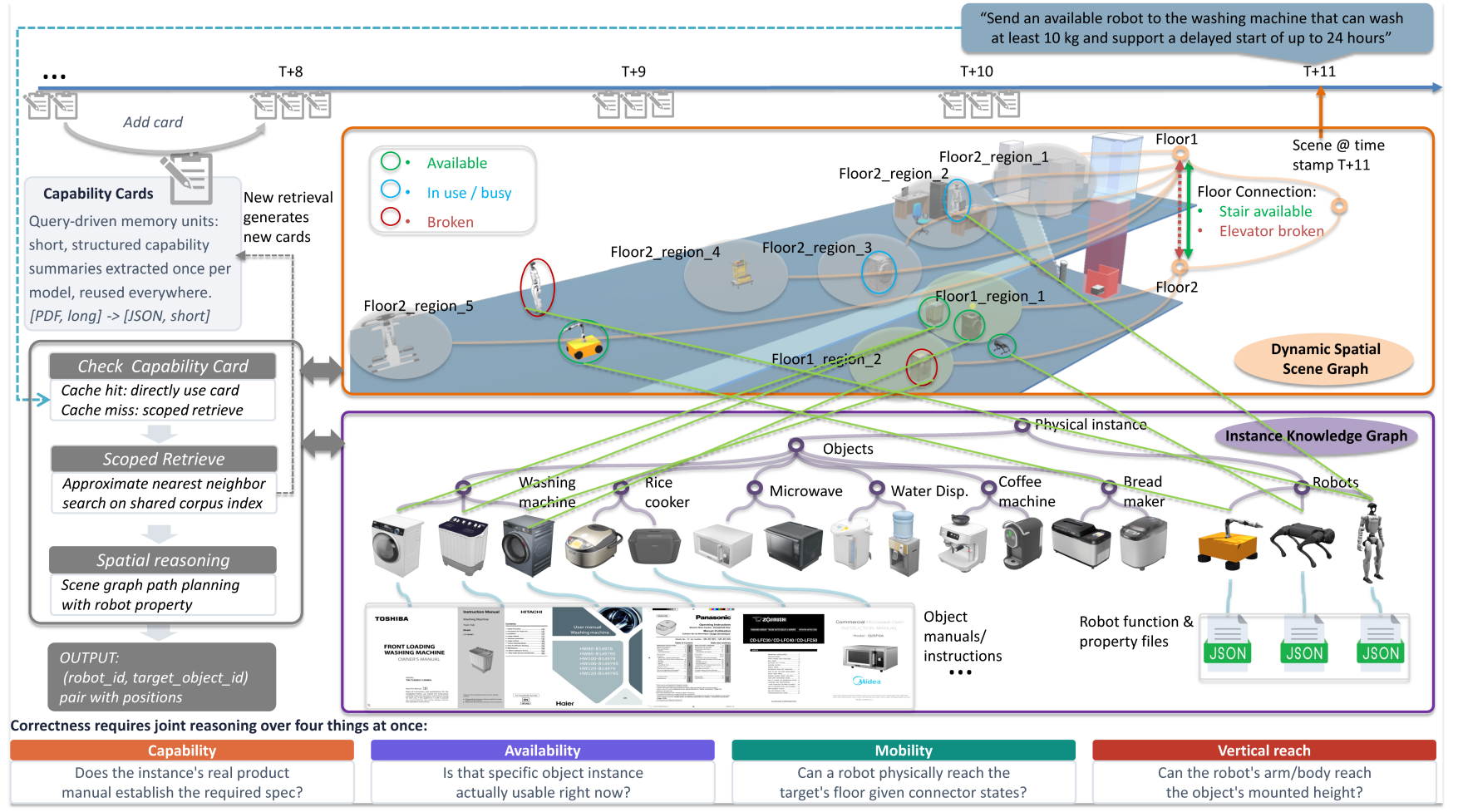}
\caption{Overview of one validated E2W-Bench instance and the Embodied RAG++ pipeline, including the dynamic hybrid scene graph, manual-linked objects, Capability Cards, scoped retrieval, and spatial reasoning. }
\label{fig:benchmark_scene_graph}
\end{figure*}

%% file: figures/dataset_stats.tex
\begin{figure*}[t]
\centering
\begin{tikzpicture}
\begin{axis}[
  width=0.99\textwidth, height=3.5cm,
  ybar, bar width=3.4pt,
  axis line style={color=gridLine}, tick style={draw=none},
  label style={font=\scriptsize, color=inkPrimary},
  tick label style={font=\scriptsize, color=inkMuted},
  ymajorgrids, major grid style={color=gridLine, line width=0.3pt},
  axis x line*=bottom, axis y line*=left,
  xlabel={Scene index}, ylabel={Instance count},
  xtick={1,...,20}, ymin=0, ymax=17,
  enlarge x limits=0.015,
  legend style={font=\scriptsize, draw=none, fill=none, at={(0.5,1.05)}, anchor=south, legend columns=-1},
]
\addplot+[draw=inkPrimary, line width=0.3pt, fill=cNaive] coordinates {
  (1,5) (2,6) (3,5) (4,5) (5,5) (6,6) (7,5) (8,5) (9,8) (10,9)
  (11,7) (12,7) (13,5) (14,6) (15,6) (16,6) (17,4) (18,5) (19,4) (20,5)
};
\addlegendentry{Manual-attached electronics}
\addplot+[draw=inkPrimary, line width=0.3pt, fill=cGraphRAG] coordinates {
  (1,3) (2,5) (3,3) (4,7) (5,4) (6,6) (7,3) (8,5) (9,2) (10,7)
  (11,4) (12,6) (13,3) (14,5) (15,2) (16,7) (17,4) (18,6) (19,3) (20,5)
};
\addlegendentry{Robots}
\addplot+[draw=inkPrimary, line width=0.3pt, fill=cDistractor] coordinates {
  (1,10) (2,5) (3,14) (4,3) (5,10) (6,7) (7,13) (8,3) (9,10) (10,15)
  (11,8) (12,2) (13,10) (14,14) (15,6) (16,2) (17,6) (18,5) (19,13) (20,6)
};
\addlegendentry{Other objects (distractors)}
\end{axis}
\node[font=\small\bfseries, color=inkPrimary] at ([yshift=0.85cm]current axis.north)
  {(a) Electronics, robot, and other-object instances per scene (E2W-0001 -- E2W-0020)};
\end{tikzpicture}

\vspace{1pt}

\begin{minipage}[t]{0.44\textwidth}
\centering
\begin{tikzpicture}
\node[font=\small\bfseries, color=inkPrimary] at (0,2.55)
  {(b) Unique Asset library category};
\def\rad{1.05}
\def\labelrad{1.2}
\draw[fill=cNaive, draw=inkPrimary, line width=0.15pt] (0,0) -- (90.0000:\rad) arc (90.0000:111.1765:\rad) -- cycle;
\draw[fill=cGraphRAG, draw=inkPrimary, line width=0.15pt] (0,0) -- (111.1765:\rad) arc (111.1765:132.3529:\rad) -- cycle;
\draw[fill=cLightRAG, draw=inkPrimary, line width=0.15pt] (0,0) -- (132.3529:\rad) arc (132.3529:164.1176:\rad) -- cycle;
\draw[fill=cEmbRAG, draw=inkPrimary, line width=0.15pt] (0,0) -- (164.1176:\rad) arc (164.1176:185.2941:\rad) -- cycle;
\draw[fill=cOurs, draw=inkPrimary, line width=0.15pt] (0,0) -- (185.2941:\rad) arc (185.2941:206.4706:\rad) -- cycle;
\draw[fill=cCoffee, draw=inkPrimary, line width=0.15pt] (0,0) -- (206.4706:\rad) arc (206.4706:227.6471:\rad) -- cycle;
\draw[fill=cDistractor, draw=inkPrimary, line width=0.15pt] (0,0) -- (227.6471:\rad) arc (227.6471:238.2353:\rad) -- cycle;
\draw[fill=cDistractor, draw=inkPrimary, line width=0.15pt] (0,0) -- (238.2353:\rad) arc (238.2353:248.8235:\rad) -- cycle;
\draw[fill=cDistractor, draw=inkPrimary, line width=0.15pt] (0,0) -- (248.8235:\rad) arc (248.8235:259.4118:\rad) -- cycle;
\draw[fill=cDistractor, draw=inkPrimary, line width=0.15pt] (0,0) -- (259.4118:\rad) arc (259.4118:270.0000:\rad) -- cycle;
\draw[fill=cDistractor, draw=inkPrimary, line width=0.15pt] (0,0) -- (270.0000:\rad) arc (270.0000:280.5882:\rad) -- cycle;
\draw[fill=cDistractor, draw=inkPrimary, line width=0.15pt] (0,0) -- (280.5882:\rad) arc (280.5882:291.1765:\rad) -- cycle;
\draw[fill=cDistractor, draw=inkPrimary, line width=0.15pt] (0,0) -- (291.1765:\rad) arc (291.1765:301.7647:\rad) -- cycle;
\draw[fill=cDistractor, draw=inkPrimary, line width=0.15pt] (0,0) -- (301.7647:\rad) arc (301.7647:312.3529:\rad) -- cycle;
\draw[fill=cDistractor, draw=inkPrimary, line width=0.15pt] (0,0) -- (312.3529:\rad) arc (312.3529:322.9412:\rad) -- cycle;
\draw[fill=cDistractor, draw=inkPrimary, line width=0.15pt] (0,0) -- (322.9412:\rad) arc (322.9412:333.5294:\rad) -- cycle;
\draw[fill=cDistractor, draw=inkPrimary, line width=0.15pt] (0,0) -- (333.5294:\rad) arc (333.5294:344.1176:\rad) -- cycle;
\draw[fill=cDistractor, draw=inkPrimary, line width=0.15pt] (0,0) -- (344.1176:\rad) arc (344.1176:354.7059:\rad) -- cycle;
\draw[fill=cDistractor, draw=inkPrimary, line width=0.15pt] (0,0) -- (354.7059:\rad) arc (354.7059:365.2941:\rad) -- cycle;
\draw[fill=cDistractor, draw=inkPrimary, line width=0.15pt] (0,0) -- (365.2941:\rad) arc (365.2941:375.8824:\rad) -- cycle;
\draw[fill=cDistractor, draw=inkPrimary, line width=0.15pt] (0,0) -- (375.8824:\rad) arc (375.8824:386.4706:\rad) -- cycle;
\draw[fill=cDistractor, draw=inkPrimary, line width=0.15pt] (0,0) -- (386.4706:\rad) arc (386.4706:397.0588:\rad) -- cycle;
\draw[fill=cDistractor, draw=inkPrimary, line width=0.15pt] (0,0) -- (397.0588:\rad) arc (397.0588:407.6471:\rad) -- cycle;
\draw[fill=cDistractor, draw=inkPrimary, line width=0.15pt] (0,0) -- (407.6471:\rad) arc (407.6471:418.2353:\rad) -- cycle;
\draw[fill=cDistractor, draw=inkPrimary, line width=0.15pt] (0,0) -- (418.2353:\rad) arc (418.2353:428.8235:\rad) -- cycle;
\draw[fill=cDistractor, draw=inkPrimary, line width=0.15pt] (0,0) -- (428.8235:\rad) arc (428.8235:439.4118:\rad) -- cycle;
\draw[fill=cDistractor, draw=inkPrimary, line width=0.15pt] (0,0) -- (439.4118:\rad) arc (439.4118:450.0000:\rad) -- cycle;
\draw[gridLine, line width=0.15pt] (100.5882:\rad) -- (100.5882:\labelrad);
\node[font=\tiny\bfseries, anchor=east, rotate=280.5882, inner sep=1.5pt] at (100.5882:\labelrad) {Microwave (2)};
\draw[gridLine, line width=0.15pt] (121.7647:\rad) -- (121.7647:\labelrad);
\node[font=\tiny\bfseries, anchor=east, rotate=301.7647, inner sep=1.5pt] at (121.7647:\labelrad) {Rice Cooker (2)};
\draw[gridLine, line width=0.15pt] (148.2353:\rad) -- (148.2353:\labelrad);
\node[font=\tiny\bfseries, anchor=east, rotate=328.2353, inner sep=1.5pt] at (148.2353:\labelrad) {Washing Machine (3)};
\draw[gridLine, line width=0.15pt] (174.7059:\rad) -- (174.7059:\labelrad);
\node[font=\tiny\bfseries, anchor=east, rotate=354.7059, inner sep=1.5pt] at (174.7059:\labelrad) {Water Disp. (2)};
\draw[gridLine, line width=0.15pt] (195.8824:\rad) -- (195.8824:\labelrad);
\node[font=\tiny\bfseries, anchor=east, rotate=375.8824, inner sep=1.5pt] at (195.8824:\labelrad) {Bread Maker (2)};
\draw[gridLine, line width=0.15pt] (217.0588:\rad) -- (217.0588:\labelrad);
\node[font=\tiny\bfseries, anchor=east, rotate=397.0588, inner sep=1.5pt] at (217.0588:\labelrad) {Coffee Machine (2)};
\draw[gridLine, line width=0.15pt] (232.9412:\rad) -- (232.9412:\labelrad);
\node[font=\tiny, anchor=east, rotate=412.9412, inner sep=1.5pt] at (232.9412:\labelrad) {Sofa};
\draw[gridLine, line width=0.15pt] (243.5294:\rad) -- (243.5294:\labelrad);
\node[font=\tiny, anchor=east, rotate=423.5294, inner sep=1.5pt] at (243.5294:\labelrad) {Cleaning Cart};
\draw[gridLine, line width=0.15pt] (254.1176:\rad) -- (254.1176:\labelrad);
\node[font=\tiny, anchor=east, rotate=434.1176, inner sep=1.5pt] at (254.1176:\labelrad) {Wall Mirror};
\draw[gridLine, line width=0.15pt] (264.7059:\rad) -- (264.7059:\labelrad);
\node[font=\tiny, anchor=east, rotate=444.7059, inner sep=1.5pt] at (264.7059:\labelrad) {Wardrobe};
\draw[gridLine, line width=0.15pt] (275.2941:\rad) -- (275.2941:\labelrad);
\node[font=\tiny, anchor=west, rotate=275.2941, inner sep=1.5pt] at (275.2941:\labelrad) {Printer};
\draw[gridLine, line width=0.15pt] (285.8824:\rad) -- (285.8824:\labelrad);
\node[font=\tiny, anchor=west, rotate=285.8824, inner sep=1.5pt] at (285.8824:\labelrad) {Exercise Bike};
\draw[gridLine, line width=0.15pt] (296.4706:\rad) -- (296.4706:\labelrad);
\node[font=\tiny, anchor=west, rotate=296.4706, inner sep=1.5pt] at (296.4706:\labelrad) {Trash Bin};
\draw[gridLine, line width=0.15pt] (307.0588:\rad) -- (307.0588:\labelrad);
\node[font=\tiny, anchor=west, rotate=307.0588, inner sep=1.5pt] at (307.0588:\labelrad) {Reception Desk};
\draw[gridLine, line width=0.15pt] (317.6471:\rad) -- (317.6471:\labelrad);
\node[font=\tiny, anchor=west, rotate=317.6471, inner sep=1.5pt] at (317.6471:\labelrad) {Floor Lamp};
\draw[gridLine, line width=0.15pt] (328.2353:\rad) -- (328.2353:\labelrad);
\node[font=\tiny, anchor=west, rotate=328.2353, inner sep=1.5pt] at (328.2353:\labelrad) {Fire Extinguisher};
\draw[gridLine, line width=0.15pt] (338.8235:\rad) -- (338.8235:\labelrad);
\node[font=\tiny, anchor=west, rotate=338.8235, inner sep=1.5pt] at (338.8235:\labelrad) {Treadmill};
\draw[gridLine, line width=0.15pt] (349.4118:\rad) -- (349.4118:\labelrad);
\node[font=\tiny, anchor=west, rotate=349.4118, inner sep=1.5pt] at (349.4118:\labelrad) {Whiteboard};
\draw[gridLine, line width=0.15pt] (360.0000:\rad) -- (360.0000:\labelrad);
\node[font=\tiny, anchor=west, rotate=360.0000, inner sep=1.5pt] at (360.0000:\labelrad) {Vending Machine};
\draw[gridLine, line width=0.15pt] (370.5882:\rad) -- (370.5882:\labelrad);
\node[font=\tiny, anchor=west, rotate=370.5882, inner sep=1.5pt] at (370.5882:\labelrad) {Dining Table};
\draw[gridLine, line width=0.15pt] (381.1765:\rad) -- (381.1765:\labelrad);
\node[font=\tiny, anchor=west, rotate=381.1765, inner sep=1.5pt] at (381.1765:\labelrad) {Filing Cabinet};
\draw[gridLine, line width=0.15pt] (391.7647:\rad) -- (391.7647:\labelrad);
\node[font=\tiny, anchor=west, rotate=391.7647, inner sep=1.5pt] at (391.7647:\labelrad) {Office Chair};
\draw[gridLine, line width=0.15pt] (402.3529:\rad) -- (402.3529:\labelrad);
\node[font=\tiny, anchor=west, rotate=402.3529, inner sep=1.5pt] at (402.3529:\labelrad) {Server Rack};
\draw[gridLine, line width=0.15pt] (412.9412:\rad) -- (412.9412:\labelrad);
\node[font=\tiny, anchor=west, rotate=412.9412, inner sep=1.5pt] at (412.9412:\labelrad) {Bulletin Board};
\draw[gridLine, line width=0.15pt] (423.5294:\rad) -- (423.5294:\labelrad);
\node[font=\tiny, anchor=west, rotate=423.5294, inner sep=1.5pt] at (423.5294:\labelrad) {Bookshelf};
\draw[gridLine, line width=0.15pt] (434.1176:\rad) -- (434.1176:\labelrad);
\node[font=\tiny, anchor=west, rotate=434.1176, inner sep=1.5pt] at (434.1176:\labelrad) {Coffee Table};
\draw[gridLine, line width=0.15pt] (444.7059:\rad) -- (444.7059:\labelrad);
\node[font=\tiny, anchor=west, rotate=444.7059, inner sep=1.5pt] at (444.7059:\labelrad) {Potted Plant};
\end{tikzpicture}
\end{minipage}%
\hfill%
\begin{minipage}[t]{0.5\textwidth}
\centering
\begin{tikzpicture}
\begin{axis}[
  width=\textwidth, height=4.3cm,
  ybar, bar width=13pt, bar shift=0pt,
  enlarge x limits=0.14,
  axis line style={color=gridLine}, tick style={draw=none},
  label style={font=\scriptsize, color=inkPrimary},
  tick label style={font=\scriptsize, color=inkMuted},
  title style={font=\small\bfseries, color=inkPrimary, yshift=-1pt},
  title={(c) Manual pages by category, full corpus}, ylabel={Pages}, ymax=340,
  ymajorgrids, major grid style={color=gridLine, line width=0.3pt},
  axis x line*=bottom, axis y line*=left,
  ymin=0,
  x tick label style={font=\scriptsize, rotate=30, anchor=east},
  nodes near coords,
  every node near coord/.append style={font=\tiny, color=inkPrimary},
  symbolic x coords={Microwave,RiceCooker,WashingMachine,WaterDisp,BreadMaker,Coffee},
  xtick={Microwave,RiceCooker,WashingMachine,WaterDisp,BreadMaker,Coffee},
  xticklabels={Microwave, Rice Cooker, Washing Machine, Water Disp., Bread Maker, Coffee Machine},
]
\addplot+[draw=inkPrimary, line width=0.3pt, fill=cNaive] coordinates {(Microwave,169)};
\addplot+[draw=inkPrimary, line width=0.3pt, fill=cGraphRAG] coordinates {(RiceCooker,50)};
\addplot+[draw=inkPrimary, line width=0.3pt, fill=cLightRAG] coordinates {(WashingMachine,118)};
\addplot+[draw=inkPrimary, line width=0.3pt, fill=cEmbRAG] coordinates {(WaterDisp,29)};
\addplot+[draw=inkPrimary, line width=0.3pt, fill=cOurs] coordinates {(BreadMaker,96)};
\addplot+[draw=inkPrimary, line width=0.3pt, fill=cCoffee] coordinates {(Coffee,308)};
\end{axis}
\end{tikzpicture}
\end{minipage}
\caption{E2W-Bench dataset composition, computed directly from the 20 dynamic scenes, the appliance instance library, and the manual corpus. (a) Manual-attached electronics, robot, and other-object (distractor) instance counts per scene. (b) The underlying instance category. (c) Number of pages in the manual corpus PDF files, totaling 770 pages.}
\label{fig:dataset_stats}
\end{figure*}
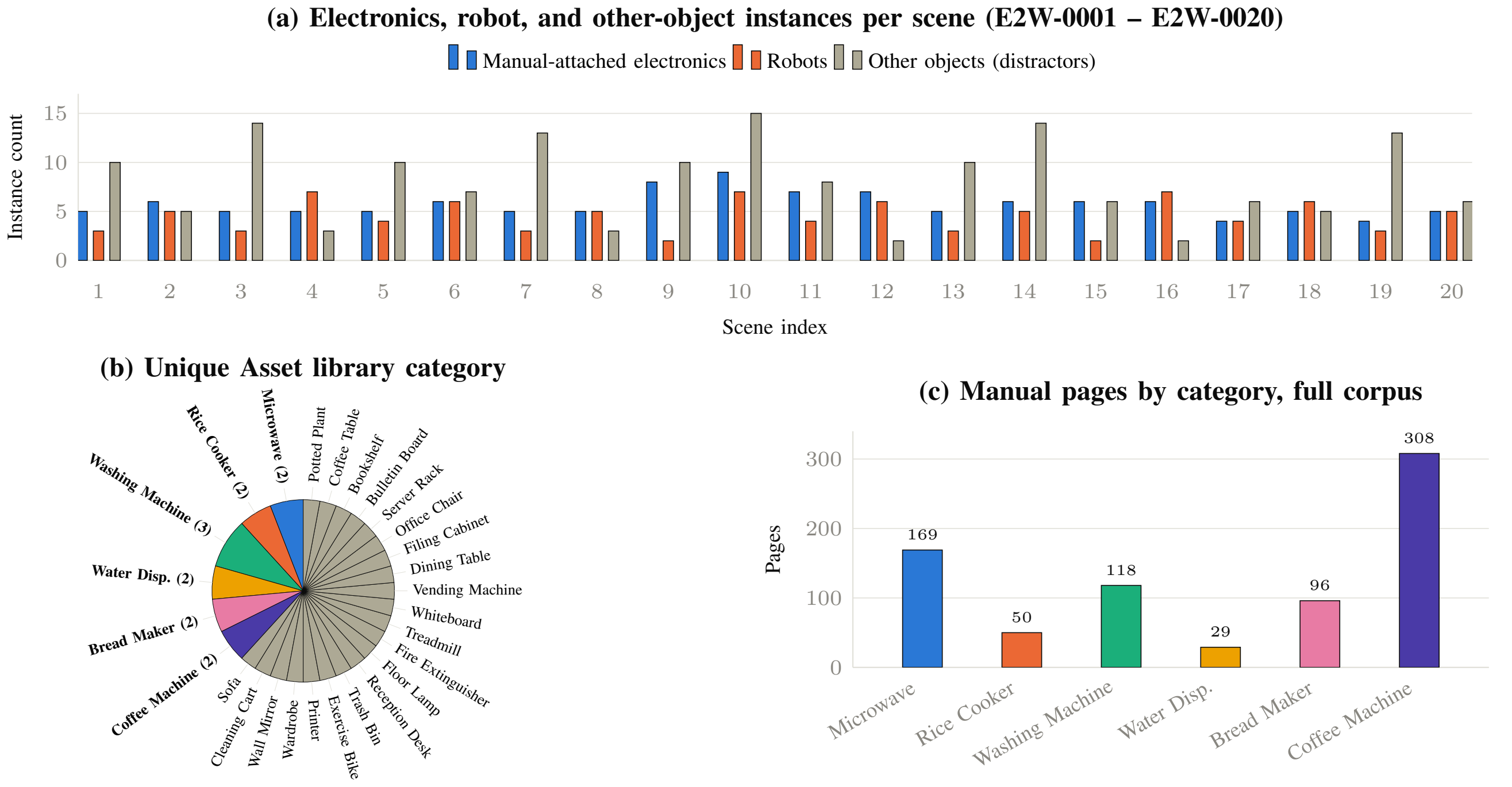

%% file: figures/efficiency_comparison.tex
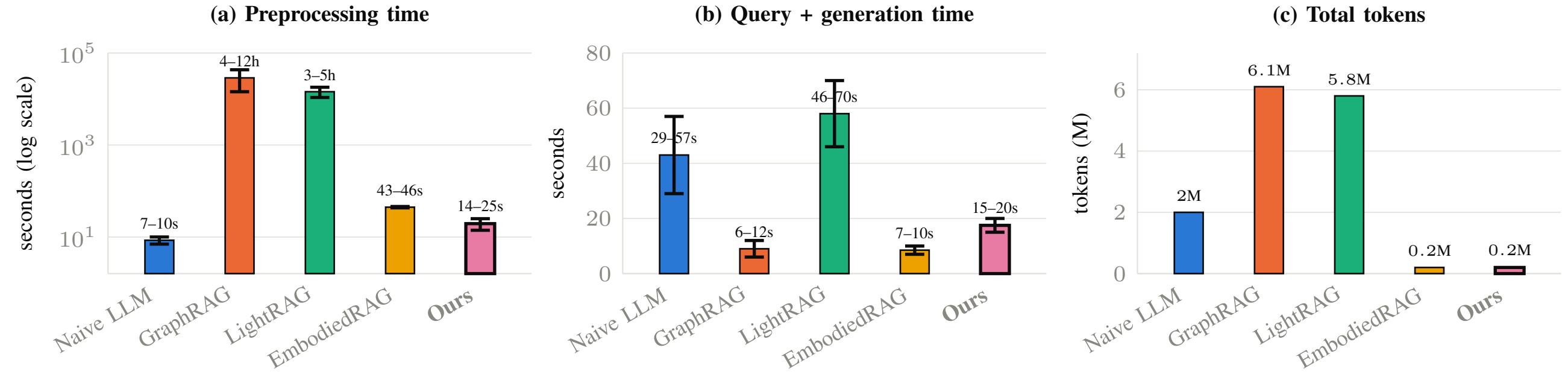
\begin{figure*}[t]
\centering
\begin{tikzpicture}
\begin{groupplot}[
  group style={group size=3 by 1, horizontal sep=1cm},
  every axis/.append style={
    width=0.35\textwidth,
    height=4cm,
    ybar,
    bar width=9pt,
    bar shift=0pt,
    enlarge x limits=0.16,
    clip=false,
    axis line style={color=gridLine},
    tick style={draw=none},
    label style={font=\scriptsize, color=inkPrimary},
    tick label style={font=\scriptsize, color=inkMuted},
    title style={font=\scriptsize\bfseries, color=inkPrimary, yshift=-1pt},
    ymajorgrids,
    major grid style={color=gridLine, line width=0.3pt},
    axis x line*=bottom,
    axis y line*=left,
    ymin=0,
    symbolic x coords={Naive,GraphRAG,LightRAG,EmbRAG,Ours},
    xtick={Naive,GraphRAG,LightRAG,EmbRAG,Ours},
    xticklabels={Naive LLM,GraphRAG,LightRAG,EmbodiedRAG,\textbf{Ours}},
    x tick label style={font=\scriptsize, rotate=30, anchor=east},
    nodes near coords,
    every node near coord/.append style={font=\tiny, color=inkPrimary},
    error bars/y dir=both,
    error bars/y explicit,
    error bars/error bar style={line width=1pt, color=inkPrimary},
    error bars/error mark options={rotate=90, mark size=3pt, line width=1pt, color=inkPrimary},
  },
]

\nextgroupplot[title={(a) Preprocessing time}, ymode=log, log origin=infty,
  ylabel={seconds (log scale)}, ymin=1.6, ymax=100000,
  point meta=explicit symbolic,
  nodes near coords={\pgfplotspointmeta}]
\addplot+[draw=inkPrimary, line width=0.5pt, fill=cNaive]
  coordinates {(Naive,8.5) +- (0,1.5) [7--10s]};
\addplot+[draw=inkPrimary, line width=0.5pt, fill=cGraphRAG]
  coordinates {(GraphRAG,28800) +- (0,14400) [4--12h]};
\addplot+[draw=inkPrimary, line width=0.5pt, fill=cLightRAG]
  coordinates {(LightRAG,14400) +- (0,3600) [3--5h]};
\addplot+[draw=inkPrimary, line width=0.5pt, fill=cEmbRAG]
  coordinates {(EmbRAG,44.5) +- (0,1.5) [43--46s]};
\addplot+[draw=inkPrimary, line width=1pt, fill=cOurs]
  coordinates {(Ours,19.5) +- (0,5.5) [14--25s]};

\nextgroupplot[title={(b) Query + generation time}, ylabel={seconds}, ymax=80,
  point meta=explicit symbolic,
  nodes near coords={\pgfplotspointmeta}]
\addplot+[draw=inkPrimary, line width=0.5pt, fill=cNaive] coordinates {(Naive,43) +- (0,14) [29--57s]};
\addplot+[draw=inkPrimary, line width=0.5pt, fill=cGraphRAG] coordinates {(GraphRAG,9) +- (0,3) [6--12s]};
\addplot+[draw=inkPrimary, line width=0.5pt, fill=cLightRAG] coordinates {(LightRAG,58) +- (0,12) [46--70s]};
\addplot+[draw=inkPrimary, line width=0.5pt, fill=cEmbRAG] coordinates {(EmbRAG,8.5) +- (0,1.5) [7--10s]};
\addplot+[draw=inkPrimary, line width=1pt, fill=cOurs] coordinates {(Ours,17.5) +- (0,2.5) [15--20s]};

\nextgroupplot[title={(c) Total tokens}, ylabel={tokens (M)}, ymax=7.2,
  nodes near coords={\pgfmathprintnumber[fixed,precision=1]{\pgfplotspointmeta}M}]
\addplot+[draw=inkPrimary, line width=0.5pt, fill=cNaive] coordinates {(Naive,2.0)};
\addplot+[draw=inkPrimary, line width=0.5pt, fill=cGraphRAG] coordinates {(GraphRAG,6.1)};
\addplot+[draw=inkPrimary, line width=0.5pt, fill=cLightRAG] coordinates {(LightRAG,5.8)};
\addplot+[draw=inkPrimary, line width=0.5pt, fill=cEmbRAG] coordinates {(EmbRAG,0.2)};
\addplot+[draw=inkPrimary, line width=1pt, fill=cOurs] coordinates {(Ours,0.2)};

\end{groupplot}
\end{tikzpicture}
\caption{Efficiency comparison across systems on the same 20 E2W-Bench scenes as Table~\ref{tab:answer_exact_match}. (a) Average one-time preprocessing time, including graph/index construction where applicable, shown on a log scale. (b) Average per-query retrieval-and-generation time. (c) Total tokens consumed across the full evaluation sweep. Bars mark the midpoint of each system's reported range; error bars span the full reported minimum--maximum.}
\label{fig:efficiency_comparison}
\end{figure*}

%% file: figures/cache_hit_rate.tex
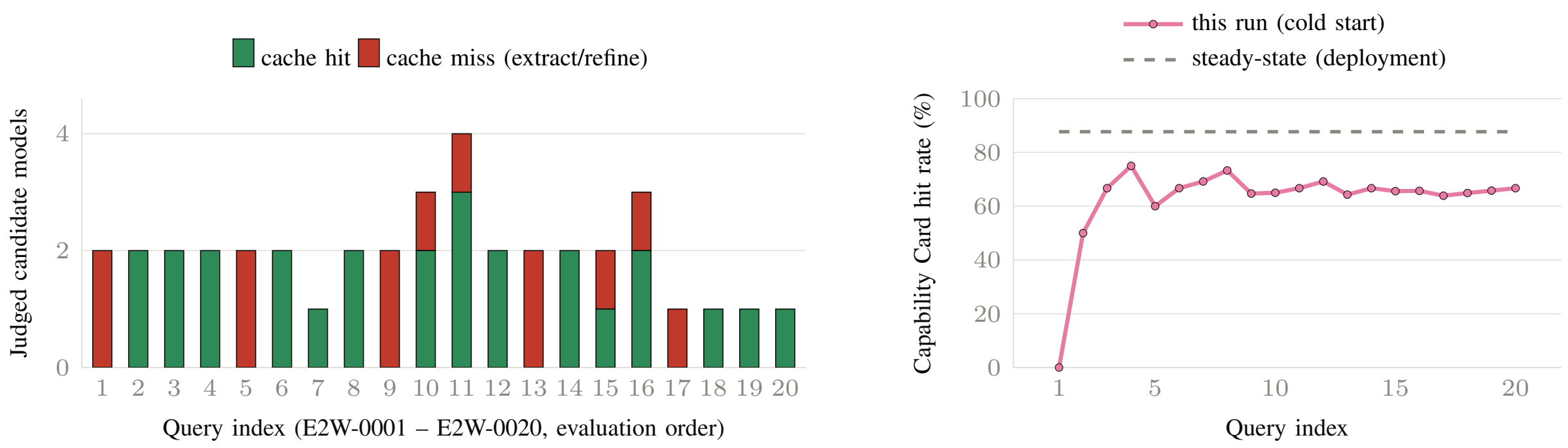
\begin{figure*}[t]
\centering
\begin{tikzpicture}
\begin{groupplot}[
  group style={group size=2 by 1, horizontal sep=2.1cm},
  every axis/.append style={
    height=4.3cm,
    axis line style={color=gridLine},
    tick style={draw=none},
    label style={font=\scriptsize, color=inkPrimary},
    tick label style={font=\scriptsize, color=inkMuted},
    ymajorgrids,
    major grid style={color=gridLine, line width=0.3pt},
    axis x line*=bottom,
    axis y line*=left,
    legend style={font=\scriptsize, draw=none, fill=none, at={(0.5,1.05)}, anchor=south, legend columns=-1},
  },
]

\nextgroupplot[
  width=0.5\textwidth,
  ybar stacked, bar width=5.5pt, enlarge x limits=0.03,
  xlabel={Query index (E2W-0001 -- E2W-0020, evaluation order)},
  ylabel={Judged candidate models},
  xtick={1,...,20}, ymin=0, ymax=4.6,
]
\addplot+[draw=inkPrimary, line width=0.3pt, fill=cCacheHit] coordinates {
  (1,0) (2,2) (3,2) (4,2) (5,0) (6,2) (7,1) (8,2) (9,0) (10,2)
  (11,3) (12,2) (13,0) (14,2) (15,1) (16,2) (17,0) (18,1) (19,1) (20,1)
};
\addlegendentry{cache hit}
\addplot+[draw=inkPrimary, line width=0.3pt, fill=cCacheMiss] coordinates {
  (1,2) (2,0) (3,0) (4,0) (5,2) (6,0) (7,0) (8,0) (9,2) (10,1)
  (11,1) (12,0) (13,2) (14,0) (15,1) (16,1) (17,1) (18,0) (19,0) (20,0)
};
\addlegendentry{cache miss (extract/refine)}
\node[font=\small\bfseries, color=inkPrimary] at ([yshift=0.95cm]current axis.north)
  {(a) Per-query cache outcomes (cold start)};

\nextgroupplot[
  width=0.4\textwidth,
  ylabel={Capability Card hit rate (\%)},
  xlabel={Query index},
  xtick={1,5,10,15,20}, ymin=0, ymax=100,
  legend style={font=\scriptsize, draw=none, fill=none, at={(0.5,1.05)}, anchor=south, legend columns=1},
  legend cell align={left},
]
\addplot+[draw=cOurs, line width=1.2pt, mark=*, mark size=1.1pt, mark options={fill=cOurs, draw=inkPrimary, line width=0.2pt}, fill=none] coordinates {
  (1,0.0) (2,50.0) (3,66.7) (4,75.0) (5,60.0) (6,66.7) (7,69.2) (8,73.3) (9,64.7) (10,65.0)
  (11,66.7) (12,69.2) (13,64.3) (14,66.7) (15,65.6) (16,65.7) (17,63.9) (18,64.9) (19,65.8) (20,66.7)
};
\addlegendentry{this run (cold start)}
\addplot+[draw=inkMuted, dashed, line width=1pt, mark=none, fill=none] coordinates {(1,87.7) (20,87.7)};
\addlegendentry{steady-state (deployment)}
\node[font=\small\bfseries, color=inkPrimary] at ([yshift=1.35cm]current axis.north)
  {(b) Cumulative hit rate over the run};

\end{groupplot}
\end{tikzpicture}
\caption{Capability Card cache analysis: (a) For each query, the number of judged candidate models whose Capability Card was already covered (hit, green) versus needed fresh extraction or refinement from the manual (miss, red). (b) Cumulative hit rate over the run (solid) against the steady-state hit rate measured over a full deployment history (dashed).}
\label{fig:cache_hit_rate}
\end{figure*}